\PassOptionsToPackage{expansion=false}{microtype}
\PassOptionsToPackage{table}{xcolor}
\RequirePackage{amsmath,amssymb,mathtools}
\documentclass[sigconf]{acmart}

\usepackage{booktabs}
\usepackage{multirow}
\usepackage{balance}
\usepackage{microtype}

\usepackage{algorithm}
\usepackage{algpseudocode}
\usepackage{amsmath}
\usepackage{xspace}

\newcommand{\sysname}{\texttt{MetaNav}\xspace}

\setcopyright{none}
\renewcommand\footnotetextcopyrightpermission[1]{}
\acmConference{}{}{}
\acmBooktitle{}
\acmYear{}
\copyrightyear{}
\acmDOI{}
\acmISBN{}
\renewcommand{\shortauthors}{}
\makeatletter
\def\@shortauthors{}
\def\@shorttitle{}
\def\ps@standardpagestyle{\ps@plain}
\def\ps@firstpagestyle{\ps@plain}
\makeatother
\fancypagestyle{standardpagestyle}{%
  \fancyhf{}%
  \renewcommand{\headrulewidth}{0pt}%
  \renewcommand{\footrulewidth}{0pt}%
  \fancyfoot[C]{\thepage}%
}
\fancypagestyle{firstpagestyle}{%
  \fancyhf{}%
  \renewcommand{\headrulewidth}{0pt}%
  \renewcommand{\footrulewidth}{0pt}%
  \fancyfoot[C]{\thepage}%
}

\begin{document}

\title{Stop Wandering: Efficient Vision-Language Navigation via Metacognitive Reasoning}

\author{Xueying Li}
\affiliation{%
  \institution{Central South University}
  \city{Changsha}
  \country{China}}

\author{Feng Lyu$^{*}$}
\affiliation{%
  \institution{Central South University}
  \city{Changsha}
  \country{China}}

\author{Hao Wu$^{*}$}
\affiliation{%
  \institution{Nanjing University}
  \city{Nanjing}
  \country{China}}

\author{Mingliu Liu}
\affiliation{%
  \institution{State Grid Hubei Electric Power Research Institute}
  \city{Wuhan}
  \country{China}}

\author{Jia-Nan Liu}
\affiliation{%
  \institution{Dongguan University of Technology}
  \city{Dongguan}
  \country{China}}

\author{Guozi Liu}
\affiliation{%
  \institution{State Grid Hubei Electric Power Research Institute}
  \city{Wuhan}
  \country{China}}

\begin{abstract} 
Training-free Vision-Language Navigation (VLN) agents powered by foundation models can follow instructions and explore 3D environments. 
However, existing approaches rely on greedy frontier selection and passive spatial memory, leading to inefficient behaviors such as local oscillation and redundant revisiting. 
We argue that this stems from a lack of metacognitive capabilities: the agent cannot monitor its exploration progress, diagnose strategy failures, or adapt accordingly. 
To address this, we propose \sysname, a metacognitive navigation agent integrating spatial memory, history-aware planning, and reflective correction. 
Spatial memory builds a persistent 3D semantic map. History-aware planning penalizes revisiting to improve efficiency. Reflective correction detects stagnation and uses an LLM to generate corrective rules that guide future frontier selection. 
Experiments on GOAT-Bench, HM3D-OVON, and A-EQA show that \sysname achieves state-of-the-art performance while reducing VLM queries by 20.7\%, demonstrating that metacognitive reasoning significantly improves robustness and efficiency. 
For access to the code, please contact xueyingli@csu.deu.cn.
\end{abstract}



\maketitle
\fancypagestyle{plain}{%
  \fancyhf{}%
  \renewcommand{\headrulewidth}{0pt}%
  \renewcommand{\footrulewidth}{0pt}%
  \fancyfoot[C]{\thepage}%
}
\pagestyle{plain}
\thispagestyle{plain}
\begingroup
\renewcommand\thefootnote{}
\footnotetext{$^{*}$ Corresponding authors.}
\endgroup

\section{Introduction}


\begin{figure*}[t]
    \centering
    \includegraphics[width=\textwidth]{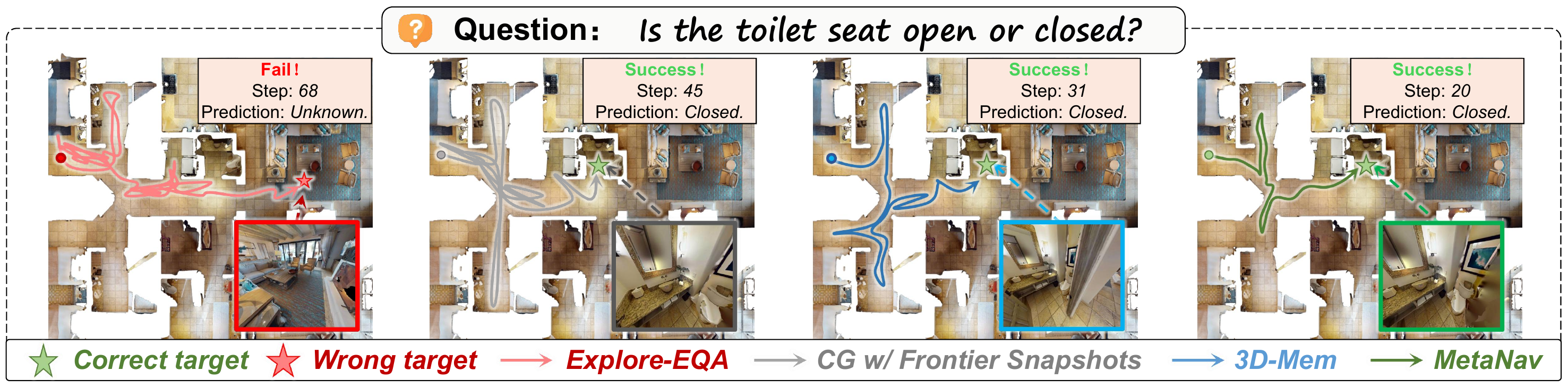}
    \caption{Qualitative trajectory comparison. Baselines suffer from local oscillation or failure when trapped by spatial ambiguities; \sysname leverages episodic reflection to break deadlocks and generate efficient paths.}
    \label{fig:fig2}
\end{figure*}

\begin{sloppypar}
Vision-Language Navigation (VLN) tasks require agents to interpret natural language instructions, perceive surrounding environments, and execute sequential decisions to reach specified targets~\cite{majumdar2022zson,yokoyama2024vlfm,kuang2024openfmnav,long2024instructnav}. 
This problem integrates perception, language understanding, spatial reasoning, and long-horizon planning, making it a challenging benchmark for embodied intelligence~\cite{zhou2026u}.

Recent advances in large-scale vision-language models (VLMs) and large language models (LLMs) have significantly improved the semantic reasoning capabilities of embodied agents. 
Leveraging these foundation models, several recent training-free approaches perform instruction grounding and frontier selection~\cite{ziliotto2025tango,zhang2025navfom,zhang2025apexnav}. 
By combining semantic perception with frontier-based exploration, these methods demonstrate promising generalization ability in previously unseen environments. 
To further improve performance, subsequent work incorporates richer global context, such as topological maps, scene captions, or retrieved historical egocentric observations~\cite{li2025flona,pan2024langnav,zhang2025mem2ego}.

Despite these advances, existing agents frequently exhibit inefficient exploration behaviors in complex environments. 
Agents may become trapped in local oscillations, repeatedly revisiting previously explored regions while failing to discover new frontiers (e.g., CG w/ Frontier Snapshots~\cite{gu2024conceptgraphs} and 3D-Mem~\cite{yang20253d} in Figure~\ref{fig:fig2}). 
In other cases, they are distracted by semantically plausible but irrelevant objects, leading to unnecessary detours (e.g., Explore-EQA~\cite{ren2024explore} in Figure~\ref{fig:fig2}). 
These behaviors result in redundant navigation trajectories, excessive model queries, and reduced task success rates. Moreover, querying large VLMs at every step incurs substantial latency, which further limits practical deployment in real-time embodied systems~\cite{ren2024explore,gu2024conceptgraphs,yang20253d,ding2025adanav,liang2024resource}.

A key reason for these failures is that most existing VLN agents rely on greedy decision policies guided by instantaneous perception or static spatial memory~\cite{wang2023gridmm,pan2024planning}.
While spatial memory records explored regions, it does not allow agents to evaluate whether their current exploration strategy is effective. As a result, agents cannot recognize when progress has stalled or when exploration has become inefficient.

We argue that effective navigation requires \textit{metacognitive capabilities}. In human exploration, individuals continuously monitor their progress, recognize when they are stuck, and revise their strategies based on past experience. This process involves reflecting on exploration history and adapting future decisions accordingly. In contrast, most existing embodied navigation systems lack mechanisms for self-monitoring and strategy correction.

To address this limitation, we propose \sysname, a metacognitive navigation framework that enables VLN agents to monitor exploration progress and adapt their strategies through reflective reasoning. The core idea is to integrate spatial perception, experience-aware planning, and reflective correction within a unified navigation architecture.
Specifically, \sysname consists of three components.
\textit{First}, a spatial memory module constructs a persistent 3D semantic map that aggregates visual observations across viewpoints, providing consistent environmental grounding for navigation decisions.
\textit{Second}, a history-aware planning module incorporates trajectory history into frontier evaluation, enabling the agent to assess exploration progress and avoid redundant revisits to previously explored regions.
\textit{Third}, a reflective correction module detects exploration stagnation and triggers reflective reasoning using an LLM to generate corrective rules that guide subsequent frontier selection.
Together, these components form a metacognitive navigation loop that allows agents to perceive the environment, evaluate exploration progress, and adapt navigation strategies when failures occur. 

We evaluate \sysname on three benchmarks: GOAT-Bench for lifelong navigation, HM3D-OVON for open-vocabulary object navigation, and A-EQA for embodied question answering. 
\sysname achieves 71.4\% success rate and 51.8\% SPL on GOAT-Bench, outperforming both training-free and supervised baselines. 
On A-EQA, it achieves 58.3\% LLM-Match, surpassing prior methods by 5.7\%. 
Ablation studies confirm the contribution of each design, and efficiency analysis shows that the replanning interval reduces VLM queries by 20.7\% compared to step-wise baselines.
Qualitative results demonstrate that \sysname successfully avoids local oscillations and generates efficient paths to targets.

Our contributions are threefold:
\begin{itemize}
    \item We propose \sysname, a metacognitive navigation framework that equips vision-language agents with the ability to monitor exploration progress, diagnose stagnation, and adapt strategies through reflective reasoning.
    
    \item \sysname integrates experience-aware planning and LLM-based reflective correction, where the agent adapts future frontier selection based on past failures, requiring no task-specific fine-tuning.
    
    \item \sysname achieves state-of-the-art performance on GOAT-Bench, HM3D-OVON, and A-EQA, while reducing VLM queries by 20.7\%. Extensive ablations validate the effectiveness of each component.
\end{itemize}
\end{sloppypar}

\section{Related Work}\label{sec:related}

\begin{sloppypar}
\textbf{Zero-Shot Vision-Language Navigation.} 
VLMs serve as the primary backbone for zero-shot embodied exploration, including object navigation and embodied question answering~\cite{majumdar2022zson,yokoyama2024vlfm,kuang2024openfmnav,long2024instructnav,das2018embodied,majumdar2024openeqa,ren2024explore}. 
Existing methods fall into three categories. 
The first uses compact trajectory encoders or lightweight state summaries for short-horizon context~\cite{pashevich2021episodic,lin2022multimodal}. 
The second introduces structured memory, such as topological graphs or semantic grid maps, for long-horizon frontier scoring~\cite{wang2021ssm,wang2023gridmm,yu2023l3mvn,yang20253d,chen2024mapgpt,sakamoto2024mapeqa}. 
The third incorporates semantic priors from foundation models via language-centric representations or adaptive fusion~\cite{pan2024langnav,li2025flona,zhang2025apexnav}. 
Despite these advances, most systems still rely on greedy semantic scoring and do not exploit failure history~\cite{krantz2023iterative,song2024longhorizon,zheng2023esceme,pan2024planning,xu2025dream,zhai2025memory}. 
These methods suffer from local oscillation in complex environments. 
Our \sysname addresses this by integrating spatial memory, history-aware planning, and metacognitive reflection.

\textbf{3D Scene Memory and Representation.} 
Embodied agents require robust representations to store and query environmental knowledge. 
Dense neural fields~\cite{xie2022neural} have been extended to object-centric and semantic 3D structures: ConceptGraphs~\cite{gu2024conceptgraphs}, GraphEQA~\cite{saxena2024grapheqa}, and KARMA~\cite{wang2025karma} organize scenes into nodes and edges, while 3D-Mem~\cite{yang20253d} introduces memory snapshots for holistic visual context. 
Further studies show that modeling geometry, semantics, and relational history benefits long-horizon navigation~\cite{wang2021ssm,singh2023scenegraph,liu2024volumetric,wang2024g3dlf,anwar2025remembr,xie2024embodied,zhang2025mem2ego,lyu2025auto}. 
Existing representations focus on maintaining a static or passively updated map, underemphasizing the temporal dynamics of exploration—such as records of past actions, outcomes, and failures~\cite{pan2024planning,xu2025dream}. 
As a result, these methods lack episodic awareness, making agents prone to repeating mistakes.
Our work introduces an episodic buffer coupled with metacognitive reflection to diagnose and correct exploration failures.

\textbf{Reflection in Embodied Agents.} 
LLMs have been applied to robotics planning via Chain-of-Thought~\cite{stechly2024chain} and ReAct~\cite{yao2022react}. 
Frameworks such as LLM-Planner~\cite{song2023llmplanner}, SayPlan~\cite{rana2023sayplan}, and SayNav~\cite{rajvanshi2024saynav} use language models to ground environmental states and update high-level plans. 
Octopus~\cite{yang2024octopus} couples embodied reasoning with executable action programs, while others deploy LLM-based agents for continuous VLN~\cite{li2024cog,ginting2025enter}. 
Nevertheless, these frameworks operate in an open loop or focus on high-level task scheduling rather than spatial error correction. 
They do not equip agents with metacognitive abilities to introspect on why a navigation attempt failed, or to resolve conflicts between semantic intuition and geometric reality (e.g., dead-ends or local minima). 
Our work develops a metacognitive reflection mechanism that monitors for exploration stagnation, invokes an LLM to analyze episodic history when the agent is stuck, and generates corrective strategies for self-correction.
\end{sloppypar}

\begin{figure*}[t]
    \centering
    \includegraphics[width=\textwidth]{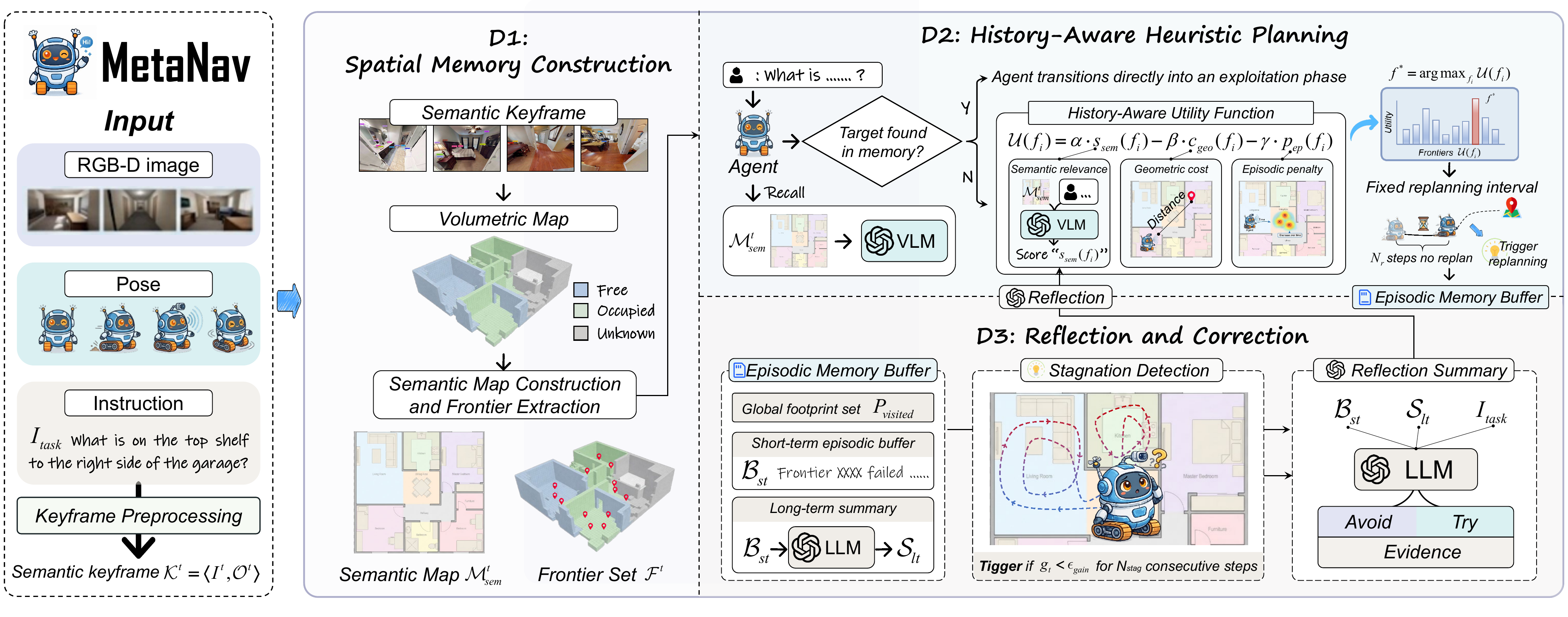}
    \caption{System overview of \sysname. Spatial Memory Construction (D1) builds a persistent 3D semantic map and extracts frontiers from RGB-D input. History-Aware Heuristic Planning (D2) selects a frontier via a utility function combining semantic relevance, geometric cost, and episodic penalty, then executes it for a fixed replanning interval. Reflection and Correction (D3) maintains episodic memory, detects stagnation via information gain, and invokes an LLM to inject corrective rules into D2. Arrows indicate data flow; dashed arrows denote LLM/VLM queries.}
    \label{fig:fig3}
\end{figure*}

\section{System Design}\label{sec:method}

\subsection{Overview}
\label{subsec:overview}

Zero-shot navigation agents often suffer from local oscillation and redundant revisiting in complex environments. 
A natural solution is to incorporate historical trajectory memory so that the agent can avoid repeating past failures. However, existing approaches either lack persistent memory or use it only as a passive map for querying, failing to diagnose failures and correct future behavior.

To bridge this gap, we identify three requirements:
\textbf{R1.} Continuous spatial mapping: the agent should dynamically build a 3D representation of its environment in real time.
\textbf{R2.} History-aware avoidance: the agent should leverage its past trajectory to suppress redundant revisiting without incurring excessive inference overhead.
\textbf{R3.} Reflective error correction: when progress stalls, the agent should detect stagnation, analyze its historical trajectory, and generate corrective strategies that override default planning.

We propose a metacognitive navigation system, \sysname, with three key designs (Figure~\ref{fig:fig3}) that address these requirements:
\textbf{D1.} Spatial Memory Construction continuously fuses egocentric RGB-D streams into a persistent 3D semantic map, providing a stable spatial foundation for both navigation and diagnosis.
\textbf{D2.} History-Aware Heuristic Planning decouples expensive VLM semantic scoring from frequent spatial execution, enabling low-latency movement while using an episodic penalty to suppress redundant revisiting.
\textbf{D3.} Reflection and Correction maintains a structured record of past actions, detects stagnation via information gain, and invokes an LLM to analyze failures and inject corrective rules back into the planner, enabling history-informed adaptation of future decisions.

\noindent \textbf{System Workflow.} 
As illustrated in Figure~\ref{fig:fig3}, these designs form a perception-planning-reflection loop. 
The loop starts with RGB-D frames as input. 
Spatial Memory Construction (D1) builds a 3D semantic map $\mathcal{M}_{sem}^t$ and extracts frontiers $\mathcal{F}^t$. 
History-Aware Heuristic Planning (D2) takes $\mathcal{M}_{sem}^t$, $\mathcal{F}^t$, and the task instruction, selects a frontier waypoint via the scoring function, and commits to it for a fixed replanning interval of $N_{replan}$ steps, during which the agent executes low-level actions without VLM calls. 
All executed positions are logged into an episodic memory buffer.
Reflection and Correction (D3) monitors the unexplored volume derived from $\mathcal{M}_{sem}^t$ to detect stagnation. 
Upon stagnation, it retrieves recent episodes from the buffer, invokes an LLM to produce corrective rules, and injects them back into D2 to modulate future frontier scoring.

\subsection{Spatial Memory Construction}
\label{sec:leve1}

To support metacognitive navigation, the agent must maintain a persistent and spatially consistent representation of its environment. 
We construct such a representation by fusing multi-view egocentric RGB-D streams into a 3D semantic map and extracting exploration frontiers. 
The process consists of the following steps.

\textbf{Keyframes Preprocessing.} 
To capture high-level semantic information from raw sensory input, we first encode each RGB-D frame into a structured observation. 
Given an RGB-D observation $I^t$, we apply a frozen VLM and an open-vocabulary segmentation model (e.g., SAM~\cite{kirillov2023segment}) to obtain a set of detected objects $\mathcal{O}^t = \{o_1,\dots,o_M\}$. 
Each object $o_j$ is represented by a tuple $(c_j, s_j, p_j)$, where $c_j$ is its semantic category, $s_j$ the detection confidence, and $p_j$ its local 3D coordinates derived from depth back-projection. 
The resulting local semantic observation is denoted as $\mathcal{K}^t = \langle I^t, \mathcal{O}^t \rangle$.

\textbf{Volumetric Mapping.} 
To maintain global spatial consistency across multiple views, we integrate semantic keyframes into a unified 3D voxel grid $\mathcal{V}$ using Truncated Signed Distance Function (TSDF) fusion~\cite{curless1996volumetric}. 
Specifically, at each time step $t$, global voxel centers are transformed into the current camera coordinate system via camera extrinsics and subsequently projected onto the image plane via intrinsics. 
By computing the signed distance between the observed raw depth and the projected voxel depth, we incrementally fuse the observations using a cumulative weighted average update. 
Rather than maintaining isolated geometric point clouds, the occupancy and exploration status of the environment are directly derived from the underlying TSDF surface and its associated confidence weights. 
Consequently, the grid is dynamically partitioned into three mutually exclusive subsets: free space $\mathcal{V}_{free}^t$, occupied space $\mathcal{V}_{occ}^t$, and unknown space $\mathcal{V}_{unk}^t$. 
Here, unknown voxels lack sufficient observation, while observed voxels are classified as free or occupied based on their TSDF signs:
\begin{equation}
\mathcal{V}^t = \mathcal{V}_{free}^t \cup \mathcal{V}_{occ}^t \cup \mathcal{V}_{unk}^t .
\end{equation}
This online volumetric representation provides a geometrically consistent basis for frontier extraction and downstream navigation.

\textbf{Semantic Map Construction and Frontiers Extraction.} 
To guide exploration toward unseen areas, we extract frontiers from the volumetric map. 
A frontier is defined as the boundary between known free space and unknown space. 
Specifically, a voxel $v$ is considered a frontier if it lies in free space and has at least one adjacent unknown voxel:
\begin{equation}
\mathcal{F}^t = \{ v \in \mathcal{V}_{free}^t \mid \exists v' \in \mathcal{N}(v), v' \in \mathcal{V}_{unk}^t \},
\end{equation}
where $\mathcal{N}(v)$ denotes the spatial neighborhood of $v$. 
In parallel, we construct a persistent, object-centric semantic map $\mathcal{M}_{sem}^t$. 
For each observation, we use the 3D object proposals from 
the semantic keyframe \(\mathcal{K}^t\) and 
associate them with the existing global object library based 
on spatial proximity and visual similarity.
Matched proposals are merged to update the geometric and semantic properties of existing global objects, while unmatched ones are instantiated as new independent entities. 
Together, $\mathcal{M}_{sem}^t$ and $\mathcal{F}^t$ form the structured spatial representation that serves as the foundation for the planning module.

\subsection{History-aware Heuristic Planning}\label{sec:leve2}

Given the spatial representation from Section~\ref{sec:leve1} and a natural language instruction, the agent must decide where to move. 
A naive approach would query a VLM at every step to score candidate frontiers, incurring high latency and often leading to greedy, oscillatory behavior. 
Our design addresses these issues with two core ideas. 
First, we decouple semantic evaluation from spatial execution: the VLM is invoked only at a fixed replanning interval to assign semantic scores to frontiers, while the agent moves continuously for multiple steps without re-querying, thereby reducing latency by avoiding expensive calls on every action. 
Second, we replace greedy VLM-only selection with a lightweight utility function that combines semantic relevance, geometric cost, and an episodic penalty that repels the agent from recently visited positions, preventing repeated attraction to spurious cues and breaking local oscillation.

\textbf{Known-target Navigation.} 
To avoid unnecessary exploration when the target is already known, the system first consults the semantic map $\mathcal{M}_{sem}^t$, which aggregates object detections from the spatial memory. 
We prompt the VLM to process $\mathcal{M}_{sem}^t$ and the recent keyframes. 
If the VLM determines that the target can already be localized within the current semantic map $\mathcal{M}_{sem}^t$, the agent directly switches to a goal-reaching policy toward its estimated spatial coordinates, bypassing frontier evaluation.

\begin{sloppypar}
\textbf{History-aware Frontier Selection.} 
When the target is not present, the system evaluates candidate frontiers $\mathcal{F}^t$ using the lightweight scoring function described in our second core idea. 
Adjacent frontier voxels are 
aggregated into frontier regions, each characterized by a 
centroid position and spatial extent. The VLM is invoked only 
at a fixed replanning interval to score all candidate frontiers 
in a single batched prompt, returning normalized semantic scores 
$s_{sem}(f_i) \in [0,1]$.
We then combine this with geometric cost and an episodic penalty and define a unified utility function $\mathcal{U}(f_i)$:
\begin{equation}
\mathcal{U}(f_i) = \alpha \cdot s_{sem}(f_i) - \beta \cdot c_{geo}(f_i) - \gamma \cdot p_{ep}(f_i).
\label{eq:cost_function}
\end{equation}
Here $c_{geo}(f_i)$ is the min‑max normalized 2D distance from the current position to the frontier centroid, encouraging nearby frontiers to reduce travel cost. 
The episodic penalty $p_{ep}(f_i)$ repels the agent from recently visited locations. 
We model the visitation history as a temporally decaying Gaussian field. Let $\{p_\tau\}_{\tau=1}^{t-1}$ be the positions visited in previous steps, which are logged during execution. The penalty is defined as:
\begin{equation}
p_{ep}(f_i) = \sum_{\tau=1}^{t-1} \lambda^{t-\tau} \exp\left(-\frac{\|p_{f_i} - p_\tau\|^2}{2\sigma^2}\right),
\end{equation}
where $p_{f_i}$ is the coordinates of frontier $f_i$, $\lambda \in (0,1)$ controls temporal decay (older visits have less influence), and $\sigma$ defines the spatial radius of repulsion. 
This penalty prevents the agent from repeatedly returning to the same dead‑ends, a common failure mode of semantic planners.

\textbf{Fixed-interval Execution.} 
To reduce latency, we avoid invoking the VLM at every step. 
After selecting the frontier $f^* = \arg\max \mathcal{U}(f_i)$, the agent commits to moving toward this waypoint for a fixed number of steps $N_{replan}$. 
During this interval, the agent executes low‑level actions without re‑querying the VLM. 
The VLM is invoked again only when the replanning interval expires or when the frontier is reached. 
This mechanism reduces the number of VLM calls and produces smoother trajectories.
\end{sloppypar}

\subsection{Reflection and Correction}\label{sec:leve3}

While the heuristic planner (Section~\ref{sec:leve2}) enables low-latency navigation, it cannot diagnose or recover from complex spatial failures. 
When the agent repeatedly visits the same areas without making progress, it requires a higher-level mechanism to analyze past behavior and adjust future decisions. 
We introduce a reflection module that maintains a structured history of past actions, detects when progress stalls, and invokes an LLM to generate corrective guidance that steers subsequent frontier evaluation.

\begin{sloppypar}
\textbf{Episodic Memory.} 
To support failure diagnosis, the agent records each execution phase into an episodic memory buffer. 
To manage long-horizon navigation without overflowing the context window, the memory is structured into two tiers:A sliding window of the $K$ most recent steps. Each entry $e_{\tau}=\langle p_{\tau}, a_{\tau}, r_{\tau}\rangle$ logs the agent's position, the action taken, and the planner's decision rationale at step $\tau$. Here, $r_{\tau}$ includes the selected frontier and the corresponding justification used during replanning.
\textit{Long-term summary $\mathcal{S}_{lt}$:} When $\mathcal{B}_{st}$ reaches capacity $K$, the LLM compresses the recent traces into a summary $\mathcal{S}_{lt}$ that captures explored regions and strategy patterns, after which the short-term buffer is cleared. 
This two-tier design preserves fine-grained recent evidence for local diagnosis while retaining coarse-grained strategic patterns over long horizons. 
Detailed prompts are reported in the supplementary material.
In parallel, all executed positions are accumulated into a global footprint set $P_{visited}$, which feeds the episodic penalty $p_{ep}(f_i)$ in history-aware heuristic planning (Section~\ref{sec:leve2}). 
\end{sloppypar}

\textbf{Stagnation Detection.} 
To identify when the agent is trapped, we monitor the unexplored volume derived from the volumetric map. 
Let $\mathcal{V}_{unk}^t$ be the unexplored volume at step $t$, and define the exploration gain as $g_t = |\mathcal{V}_{unk}^{t-1}| - |\mathcal{V}_{unk}^t|$. 
Stagnation is detected if $g_t < \epsilon_{gain}$ for $N_{stag}$ consecutive steps. 
To prevent repeated reflections, a cooldown $T_{cool}$ is enforced. 
The deadlock trigger $\Delta_t$ is defined as:
\begin{equation}
    \Delta_t = \mathbb{I} \left( \sum_{\tau=t-N_{stag}}^t \mathbb{I}(g_\tau < \epsilon_{gain}) \ge N_{stag} \land (t - t_{last}) \ge T_{cool} \right),
\end{equation}
where $\mathbb{I}(\cdot)$ is the indicator function and $t_{last}$ is the timestamp of the previous reflection.

\textbf{LLM-based correction.} 
When $\Delta_t = \texttt{True}$, the agent pauses heuristic planning and invokes the LLM for error correction. 
The LLM is prompted with the instruction $I_{task}$, the long-term summary $\mathcal{S}_{lt}$, and the recent failed attempts in $\mathcal{B}_{st}$. 
It outputs a \textit{reflection summary} $\mathcal{R}_t$ with three components: \textit{Avoid} (ineffective regions or heuristics to skip), \textit{Try} (promising alternative directions), and \textit{Evidence} (the logical deduction from past failures). 
Case studies are reported in the supplementary material. 
The reflection summary $\mathcal{R}_t$ is concatenated to the prompt used by the VLM in history-aware heuristic planning, so that the ``Avoid/Try'' rules influence semantic scoring in subsequent replanning cycles. 
Algorithm~\ref{alg:reflection} summarizes this online interaction between history-aware planning and metacognitive reflection, including known-target navigation, gain-based stagnation detection, frontier scoring, and memory update.

\begin{algorithm}[t]
    \caption{Online Planning and Reflection in \sysname}
    \label{alg:reflection}
    \begin{algorithmic}[1]
    \Require Semantic Map $\mathcal{M}_{sem}^t$, Frontiers $\mathcal{F}^t$, Instruction $I_{task}$, Episodic Memory $\mathcal{B}_{st}, \mathcal{S}_{lt}$, Visited Positions $P_{visited}$, Unexplored Volume $\mathcal{V}_{unk}^t$, Current Position $p_t$, Previous Reflection Time $t_{last}$
    \Ensure Selected waypoint, executed trajectory segment, and updated memory state
    \State Initialize reflection summary $\mathcal{R} \leftarrow \emptyset$
    \For{each planning cycle $t$}
        \Statex \textcolor{gray} {Phase 1: Known-target navigation}
        \If{$\text{IsTargetLocalized}(\mathcal{M}_{sem}^t, I_{task})$}
            \State $p_{target} \leftarrow \text{GetTargetCoordinates}(\mathcal{M}_{sem}^t, I_{task})$
            \State $\tau_t \leftarrow \text{NavigateTo}(p_{target})$
            \State \textbf{return} \Comment{Switch to goal-reaching execution}
        \EndIf
        
        \Statex \textcolor{gray} {Phase 2: Stagnation detection and reflection}
        \State $\Delta_t \leftarrow \text{DetectStagnation}(\mathcal{V}_{unk}^t, t, t_{last})$ \Comment{Eq. 5}
        \If{$\Delta_t$ is \textbf{True}}
            \State $\mathcal{R}_t \leftarrow \text{GenerateReflection}(\mathcal{B}_{st}, \mathcal{S}_{lt}, I_{task})$
        \EndIf
        
        \Statex \textcolor{gray} {Phase 3: History-aware frontier selection}
        \State $\mathcal{P}_{prompt} \leftarrow \text{Concat}(I_{task}, \mathcal{M}_{sem}^t, \mathcal{B}_{st}, \mathcal{S}_{lt}, \mathcal{R}_t)$
        \For{each frontier $f_i$ in ${\mathcal{F}}^t$}
            \State $s_{sem}(f_i) \leftarrow \text{VLM\_Score}(\mathcal{P}_{prompt}, f_i)$
            \State $c_{geo}(f_i) \leftarrow \text{GeoCost}(p_t, f_i)$
            \State $p_{ep}(f_i) \leftarrow \text{ComputePenalty}(f_i, P_{visited})$ \Comment{Eq.~4}
            \State $\mathcal{U}(f_i) \leftarrow \alpha \cdot s_{sem}(f_i) - \beta \cdot c_{geo}(f_i) - \gamma \cdot p_{ep}(f_i)$ \Comment{Eq.~3}
        \EndFor
        \State $f^* \leftarrow \arg\max \mathcal{U}(f_i)$
        \State $\tau_t \leftarrow \text{Execute}(f^*, N_{replan})$ \Comment{Commit to $f^*$ for fixed steps}
        
        \Statex \textcolor{gray} {Phase 4: Memory update}
        \State $P_{visited} \leftarrow P_{visited} \cup \tau_t$
        \State $(\mathcal{B}_{st}, \mathcal{S}_{lt}) \leftarrow \text{UpdateMemory}(\mathcal{B}_{st}, \mathcal{S}_{lt}, \tau_t, f^*, \mathcal{R}_t)$
    \EndFor
    \end{algorithmic}
\end{algorithm}

\section{Experiments}\label{sec:experiments}

\subsection{Experimental Settings}
\begin{sloppypar}
\textbf{Benchmarks.} 
We evaluate on three benchmarks that cover different aspects of embodied navigation and reasoning.
\textbf{GOAT-Bench}~\cite{khanna2024goat} is a lifelong visual navigation benchmark that tests the agent's ability to locate open-vocabulary instances specified by multi-modal instructions in complex, unstructured environments.
\textbf{HM3D-OVON}~\cite{yokoyama2024hm3dovon} is a standard testbed for open-vocabulary object navigation, requiring the agent to find arbitrary target categories across diverse 3D indoor scenes.
\textbf{A-EQA}~\cite{majumdar2024openeqa} assesses embodied question answering performance, where the agent must navigate to specific viewpoints to answer questions about object states or relations in high-fidelity HM3D scenes~\cite{ramakrishnan2021habitat}.

\textbf{Evaluation Metrics.}
For GOAT-Bench and HM3D-OVON, we report Success Rate (SR) and Success weighted by Path Length (SPL). 
An episode is successful if the agent stops within 1.0 m of the target object. 
SPL weights success by path efficiency, defined as the ratio of the shortest geodesic distance to the actual path length.
For A-EQA, we use LLM-Match and LLM-SPL. 
LLM-Match employs a frozen LLM to semantically verify the consistency between the generated answer and the ground truth, accommodating linguistic variations. 
LLM-SPL weights the semantic accuracy score by the path length ratio.

\textbf{Baselines.} We compare against two categories of methods. 
\textbf{(1) Supervised methods.} 
These require environment-specific training or large-scale trajectory pre-training. 
They include Uni-NaVid~\cite{zhang2025uninavid}, OVSegDT~\cite{zemskova2025ovsegdt}, Dynam3D~\cite{wang2025dynam3d}, SenseAct-NN (Monolithic and Skill Chain) on GOAT-Bench~\cite{khanna2024goat}; BC~\cite{pomerleau1988alvinn}, DAgger~\cite{ross2011reduction}, RL~\cite{schulman2017proximal}, and DAgRL on HM3D-OVON~\cite{yokoyama2024hm3dovon}; and MTU3D~\cite{zhu2025mtu3d} and NavFoM~\cite{zhang2025navfom}.
\textbf{(2) Training-free method:} 
These leverage pre-trained foundation models without fine-tuning. 
They include heuristic semantic planners (Modular GOAT~\cite{chang2023goat}, Modular CLIP on Wheels~\cite{gadre2023cows}, VLFM~\cite{yokoyama2024vlfm}), scene graph and spatial memory approaches (ConceptGraphs~\cite{gu2024conceptgraphs}, 3D-Mem~\cite{yang20253d}), and VLM-driven exploration agents (Explore-EQA~\cite{ren2024explore}, TANGO~\cite{ziliotto2025tango}). 
For A-EQA, we additionally compare against vision-blind LLMs (LLaMA-2~\cite{touvron2023llama}, GPT-4~\cite{openai2023gpt4}) and VLMs with captioning strategies~\cite{majumdar2024openeqa,liu2024improved,yang2023dawn}.

\end{sloppypar}

\textbf{Implementation Details.}
Each episode has a maximum of 50 steps. 
The agent receives $1280 \times 1280$ RGB-D images with a $120^\circ$ horizontal field of view; the effective depth range is 1.7 m. 
The camera is positioned 1.5 m above the ground with a $30^\circ$ downward tilt. 
The maximum moving distance per step is 1.0 m.
For perception, we use YOLOv8x-World~\cite{cheng2024yolo} for open-vocabulary object detection and SAM-L~\cite{kirillov2023segment} for instance segmentation. 
GPT-4o serves as the VLM for decision-making and as the LLM for reflection. 
Visual inputs to the VLM are downsampled to $360 \times 360$ for computational efficiency.

\begin{table}[h]
    \centering
    \caption{Results on the Val Unseen split of GOAT-Bench. Best and second-best results are in \textbf{bold} and \underline{underline}.}
    \label{tab:goat_bench_performance}
    \setlength{\tabcolsep}{12pt}
    \small
    \begin{tabular}{l||cc}
        \toprule
        \multicolumn{1}{l||}{\textbf{Method}} & \textbf{SR $\uparrow$} & \textbf{SPL $\uparrow$} \\
        \midrule
        \rowcolor{gray!20} \multicolumn{3}{l}{\textit{Supervised methods}} \\
        SenseAct-NN Monolithic~\cite{khanna2024goat} & 12.3 & 6.8 \\
        SenseAct-NN Skill Chain~\cite{khanna2024goat} & 29.5 & 11.3 \\
        MTU3D~\cite{zhu2025mtu3d} & 47.2 & 27.7 \\
        \midrule
        \rowcolor{gray!20} \multicolumn{3}{l}{\textit{Training-free methods}} \\
        Modular CLIP on Wheels~\cite{gadre2023cows} & 16.1 & 10.4 \\
        Modular GOAT~\cite{chang2023goat} & 24.9 & 17.2 \\
        TANGO~\cite{ziliotto2025tango} & 32.1 & 16.5 \\
        Explore-EQA~\cite{ren2024explore} & 61.5 & 45.3 \\
        CG w/ Frontier Snapshots~\cite{gu2024conceptgraphs} & 55.0 & 37.9 \\
        3D-Mem~\cite{yang20253d} & \underline{69.1} & \underline{48.9} \\
        \midrule
        \textbf{\sysname (Ours)} & \textbf{71.4} & \textbf{51.8} \\
        \bottomrule
    \end{tabular}
\end{table}

\subsection{Overall Results}

\begin{sloppypar}
\textbf{GOAT-Bench.} 
Table~\ref{tab:goat_bench_performance} reports results on the Val Unseen split. 
\sysname achieves a success rate (SR) of 71.4\% and SPL of 51.8\%, outperforming all baselines. 
Compared to the second-best method, 3D-Mem, this represents an improvement of 2.3\% in SR and 2.9\% in SPL. 
\sysname also exceeds the best supervised method, MTU3D, by 24.2\% in SR. 
The simultaneous increase in both metrics indicates that the geometric cost and an episodic penalty function Eq.~\ref{eq:cost_function} effectively reduce redundant revisiting and local oscillation.

\begin{figure}[t]
    \centering
    \includegraphics[width=\columnwidth]{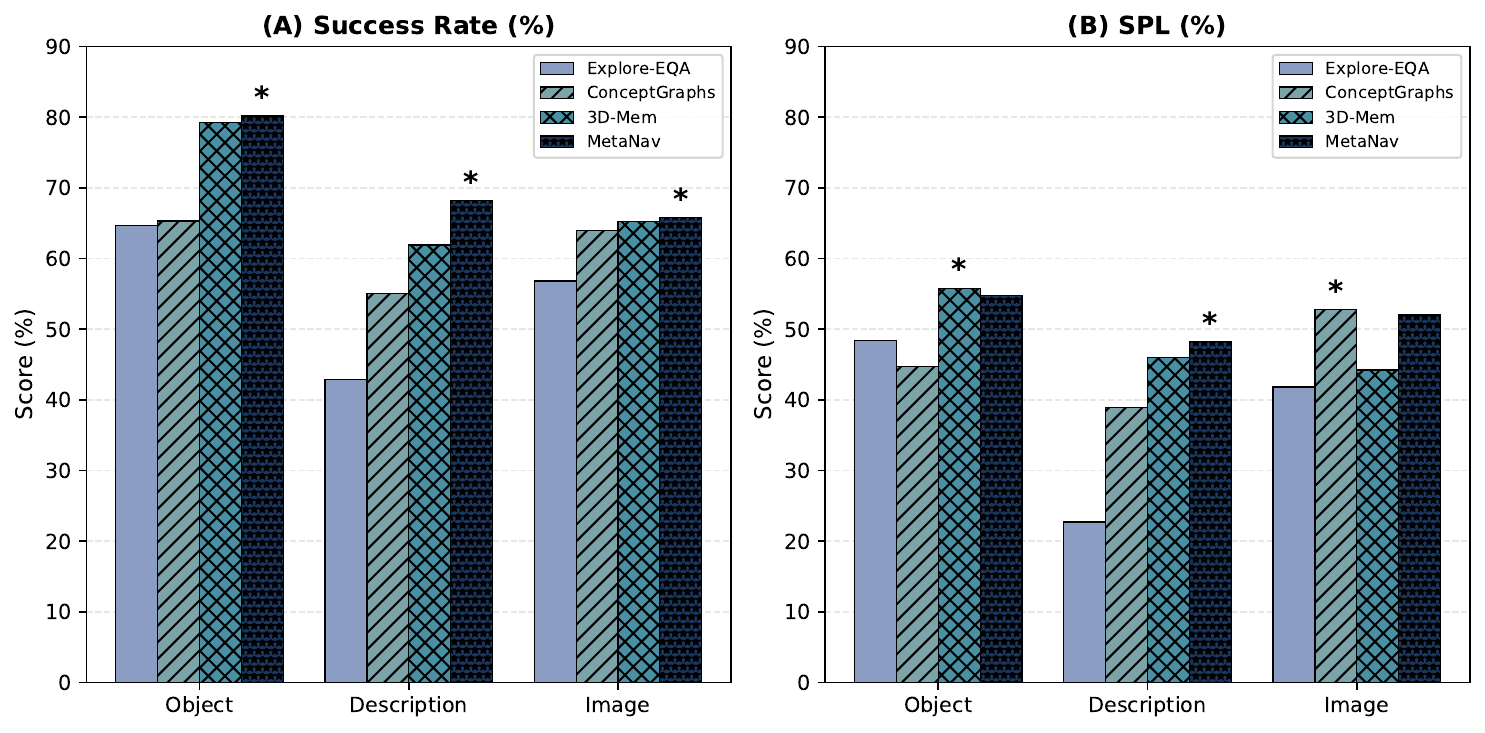}
    \caption{Performance across instruction modalities on GOAT-Bench.}
    \label{fig:goatbench}
\end{figure}

Figure~\ref{fig:goatbench} breaks down performance across three instruction modalities: Object, Description, and Image. 
\sysname achieves the best SR across all three, with a notable advantage on Description tasks. 
In SPL, \sysname maintains balanced performance across modalities, while baselines show larger fluctuations.
\end{sloppypar}

\begin{sloppypar}
\textbf{HM3D-OVON.} 
Table~\ref{tab:hm3d_ovon_performance} reports results on the Val Unseen split. 
\sysname achieves the best SR among training-free methods at 46.1\%, also surpassing all supervised baselines including the strongest, NavFoM (45.2\%), which requires large-scale trajectory pre-training. 
NavFoM achieves a higher SPL (31.9\% vs. 29.8\%), suggesting that its end-to-end trained policy is more path-efficient; however, \sysname attains higher success without any task-specific training, demonstrating the effectiveness of metacognitive exploration under a training-free setting.
\end{sloppypar}

\begin{table}[h]
    \centering
    \caption{Results on the Val Unseen split of HM3D-OVON. Best and second-best results are in \textbf{bold} and \underline{underline}.}
    \label{tab:hm3d_ovon_performance}
    \setlength{\tabcolsep}{20pt}
    \small
    \begin{tabular}{l||cc}
        \toprule
        \multicolumn{1}{l||}{\textbf{Method}} & \textbf{SR $\uparrow$} & \textbf{SPL $\uparrow$} \\
        \midrule
        \rowcolor{gray!20} \multicolumn{3}{l}{\textit{Supervised methods}} \\
        BC~\cite{pomerleau1988alvinn} & 5.4 & 1.9 \\
        DAgger~\cite{ross2011reduction} & 10.2 & 4.7 \\
        DAgRL~\cite{yokoyama2024hm3dovon} & 18.3 & 7.9 \\
        RL~\cite{schulman2017proximal} & 18.6 & 7.5 \\
        DAgRL+OD~\cite{yokoyama2024hm3dovon} & 37.1 & 19.8 \\
        Uni-NaVid~\cite{zhang2025uninavid} & 39.5 & 19.8 \\
        OVSegDT~\cite{zemskova2025ovsegdt} & 40.1 & 20.9 \\
        MTU3D~\cite{zhu2025mtu3d} & 40.8 & 12.1 \\
        Dynam3D~\cite{wang2025dynam3d} & 42.7 & 22.4 \\
        NavFoM~\cite{zhang2025navfom} & \underline{45.2} & \textbf{31.9} \\
        \midrule
        \rowcolor{gray!20} \multicolumn{3}{l}{\textit{Training-free methods}} \\
        Modular GOAT~\cite{khanna2024goat} & 24.9 & 17.2 \\
        VLFM~\cite{yokoyama2024vlfm} & 35.2 & 19.6 \\
        TANGO~\cite{ziliotto2025tango} & 35.5 & 19.5 \\
        
        \midrule
        \textbf{\sysname (Ours)} & \textbf{46.1} & \underline{29.8} \\
        \bottomrule
    \end{tabular}
\end{table}

\textbf{A-EQA.} 
Table~\ref{tab:llm_vlm_performance} reports results on embodied question answering. 
\sysname achieves LLM-Match of 58.3\% and LLM-SPL of 45.5\%, outperforming the best baseline 3D-Mem by 5.7\% and 3.5\%, respectively. 
Methods without spatial memory, such as GPT-4V, achieve only 41.8\% LLM-Match, highlighting the importance of memory and reflection for embodied question answering.

\begin{table}[h]
    \centering
    \caption{Results on A-EQA. Best and second-best results are in \textbf{bold} and \underline{underline}.}
    \label{tab:llm_vlm_performance}
    \setlength{\tabcolsep}{7pt}
    \small
    \begin{tabular}{l||cc}
        \toprule
        \multicolumn{1}{l||}{\textbf{Method}} & \textbf{LLM-Match $\uparrow$} & \textbf{LLM-SPL $\uparrow$} \\
        \midrule
        \rowcolor{gray!20} \multicolumn{3}{l}{\textit{Blind LLMs}} \\
        LLaMA-2~\cite{touvron2023llama} & 29.0 & N/A \\
        GPT-4~\cite{openai2023gpt4} & 35.5 & N/A \\
        \midrule
        \rowcolor{gray!20} \multicolumn{3}{l}{\textit{LLM with captions}} \\
        LLaMA-2 w/ LLaVA-1.5~\cite{liu2024improved} & 30.9 & 5.9 \\
        GPT-4 w/ LLaVA-1.5~\cite{liu2024improved} & 38.1 & 7.0 \\
        \midrule
        \rowcolor{gray!20} \multicolumn{3}{l}{\textit{VLMs}} \\
        GPT-4V~\cite{yang2023dawn} & 41.8 & 7.5 \\
        \midrule
        \rowcolor{gray!20} \multicolumn{3}{l}{\textit{Embodied VLM Exploration}} \\
        Explore-EQA~\cite{ren2024explore} & 46.9 & 23.4 \\
        CG w/ Frontier Snapshots~\cite{gu2024conceptgraphs} & 47.2 & 33.3 \\
        3D-Mem~\cite{yang20253d} & \underline{52.6} & 42.0 \\
        MTU3D + GPT-4o~\cite{zhu2025mtu3d} & 51.1 & \underline{42.6} \\
        \midrule
        \textbf{\sysname (Ours)} & \textbf{58.3} & \textbf{45.5} \\
        \bottomrule
    \end{tabular}
\end{table}

\subsection{Ablation Study}
\begin{sloppypar}
Table~\ref{tab:ablation} presents ablation results on GOAT-Bench Val Unseen. 
Removing reflection and correction (D3) drops SR from 71.4\% to 66.3\%, indicating that reflective error correction is essential for breaking deadlocks. 
Removing the episodic memory buffer while keeping reflection reduces SR to 69.2\%, showing that reflection without historical context is less effective. 
Removing the long-term summary further reduces SR to 68.7\%, confirming its value for long-horizon tasks.
Replacing history-aware heuristic planning (D2) with a greedy VLM-only scorer, i.e., disabling the unified scoring function $\mathcal{U}$ (Eq.\ref{eq:cost_function}), drops SR to 64.9\%, confirming that the combination of semantic relevance, geometric cost, and episodic penalty is critical for efficient exploration.
Removing spatial memory construction (D1) reduces SR to 58.6\% and SPL to 40.0\%, demonstrating that persistent spatial mapping is fundamental to avoiding redundant re-exploration.
\end{sloppypar}

\begin{table}[h]
    \caption{Ablation study on GOAT-Bench Val Unseen.}
    \label{tab:ablation}
    \setlength{\tabcolsep}{7pt}
    \centering
    \small
    \renewcommand{\arraystretch}{1.1}
    \begin{tabular}{l||cc}
    \toprule
    \textbf{Architecture Variant} & \textbf{SR $\uparrow$} & \textbf{SPL $\uparrow$} \\
    \midrule
    \textbf{\sysname (Full Architecture)} & \textbf{71.4} & \textbf{51.8} \\
    \midrule
    \rowcolor{gray!20} \multicolumn{3}{l}{\textit{Ablating D3: Reflection and Correction}} \\
    \quad \textit{w/o} Long-Term Summary & 68.7 & 49.6 \\ 
    \quad \textit{w/o} Episodic Memory & 69.2 & 49.9 \\ 
    \quad \textit{w/o} Reflection & 66.3 & 47.1 \\ 
    \midrule
    \rowcolor{gray!20} \multicolumn{3}{l}{\textit{Ablating D2: History-Aware Heuristic Planning}} \\
    \quad \textit{w/o} Unified Scoring Function (greedy VLM) & 64.9 & 48.2 \\ 
    \midrule
    \rowcolor{gray!20} \multicolumn{3}{l}{\textit{Ablating D1: Spatial Memory Construction}} \\
    \quad \textit{w/o} Spatial Memory Construction  & 58.6 & 40.0 \\ 
    \bottomrule
    \end{tabular}
\end{table}

\subsection{Sensitivity and System Efficiency Analysis}

\begin{figure}[t]
    \centering
    \includegraphics[width=\columnwidth]{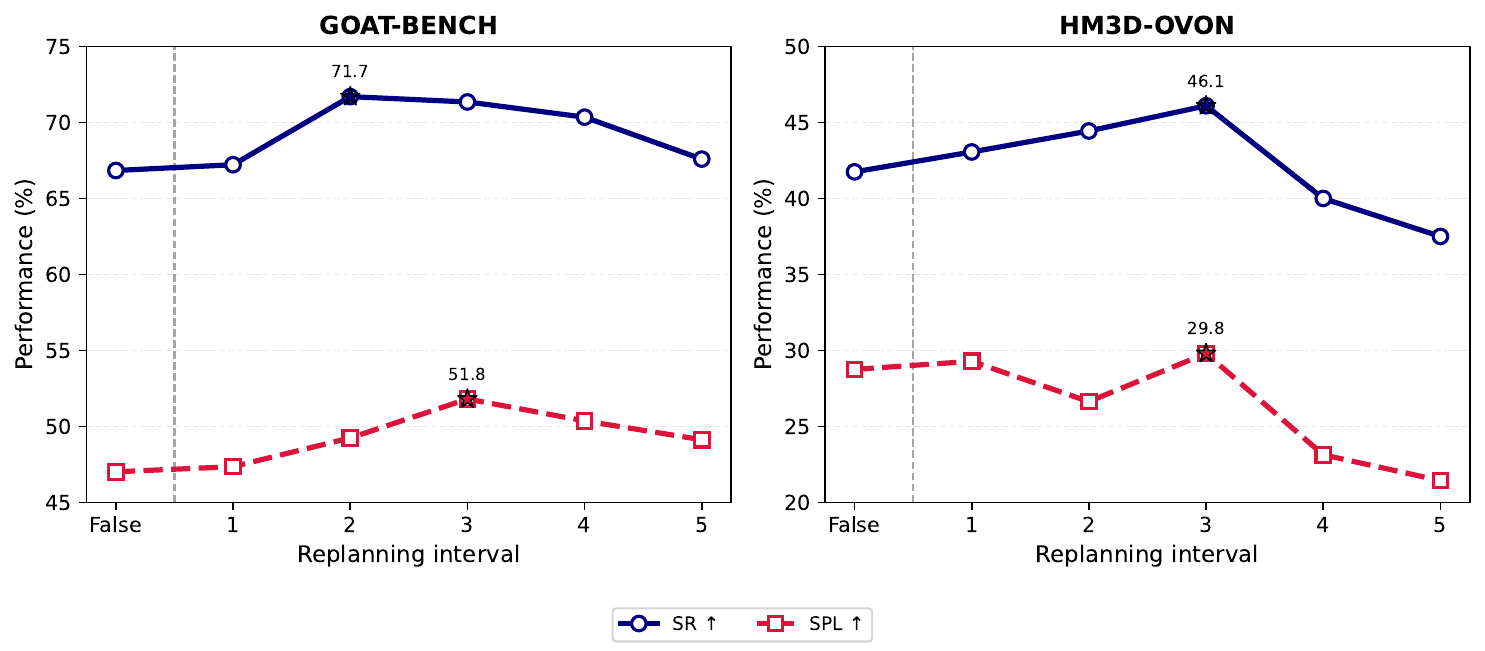}
    \caption{Effect of replanning interval on GOAT-Bench and HM3D-OVON.}
    \label{fig:steps}
\end{figure}

\textbf{Replanning Interval.} 
Figure~\ref{fig:steps} shows the effect of the replanning interval $N_{replan}$ on GOAT-Bench and HM3D-OVON. 
As the interval increases, performance initially rises and then declines. 
On GOAT-Bench, peak SR (71.7\%) occurs at $N_{replan}=2$, while peak SPL (51.8\%) occurs at $N_{replan}=3$. 
On HM3D-OVON, both metrics peak at $N_{replan}=3$, reaching 46.1\% SR and 29.8\% SPL. 
This confirms that an appropriate replanning interval prevents local jittering while avoiding over-commitment to distant frontiers.
Based on these results, we set $N_{replan}=3$ for all main experiments.

\begin{figure}[h]
    \centering
    \includegraphics[width=\columnwidth]{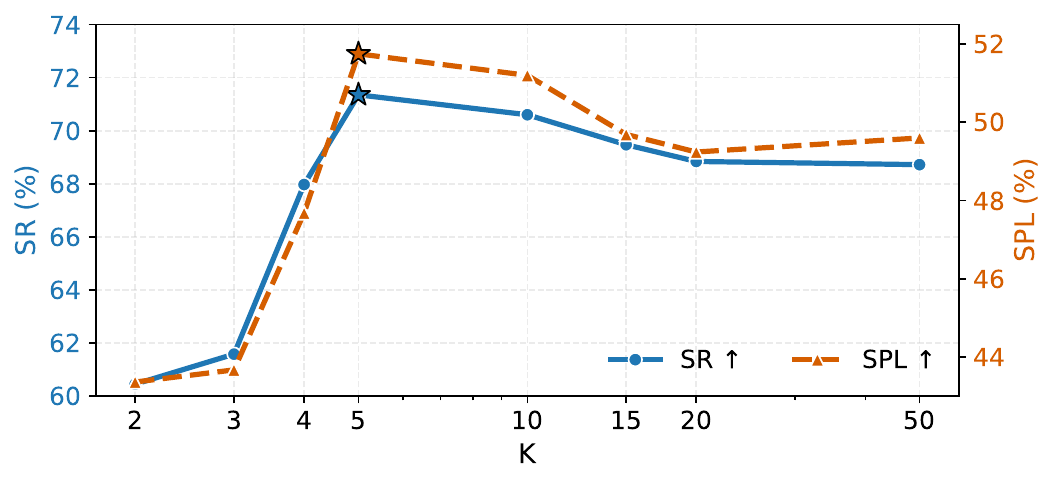}
    \caption{Effect of short-term memory capacity $K$ on GOAT-Bench.}
    \label{fig:capacity}
\end{figure}

\textbf{Short-Term Memory Capacity.} 
Figure~\ref{fig:capacity} shows the effect of the short-term buffer size $K$ on GOAT-Bench. 
Performance peaks at $K=5$ (71.35\% SR, 51.75\% SPL). 
When $K$ is too small ($K=2$), SR drops to 60.5\%, indicating that frequent compression disrupts local exploration coherence and leads to excessive context fragmentation. 
When $K$ is too large ($K=50$), SR drops to 68.7\%, suggesting that the LLM struggles to extract actionable reasoning from long sequences of raw trajectory logs. 
The optimal $K=5$ balances these trade-offs, providing enough context for error diagnosis while filtering out noisy details.

\begin{table}[h]
    \caption{Average latency of \sysname components.}
    \label{tab:latency}
    \centering
    \small
    \setlength{\tabcolsep}{8pt}
    \begin{tabular}{l||l||c}
    \toprule
    \textbf{Design} & \textbf{Component} & \textbf{Latency} \\
    \midrule
    \multirow{2}{*}{Design 1} 
    & Point Cloud Projection & 54.0 ms \\
    & Memory Fusion & 359.9 ms \\
    \midrule
    \multirow{2}{*}{Design 2}
    & Unified Scoring \& Penalty & 2.0 ms \\
    & LLM Inference & 5.40 s \\
    \midrule
    Design 3 & Reflection \& Summary & 5.75 s \\
    \bottomrule
    \end{tabular}
\end{table}

\begin{figure*}[t]
    \centering
    \includegraphics[width=\textwidth]{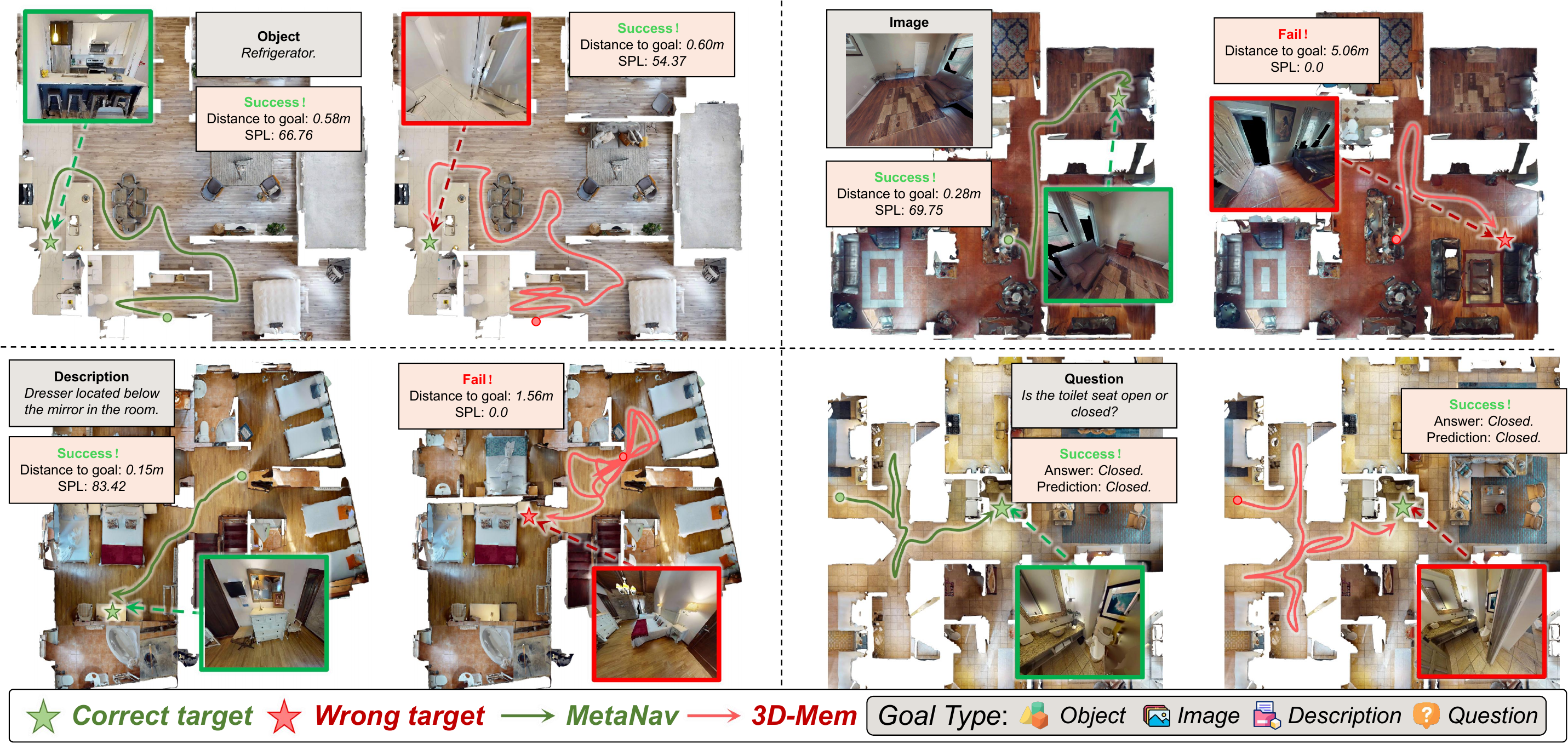}
    \caption{Trajectory comparison across four goal modalities. 3D-Mem (red) shows local oscillation and target confusion; \sysname (green) produces efficient paths.}
    \label{fig:qualitative}
\end{figure*}

\textbf{Latency.} 
Table~\ref{tab:latency} reports average latencies of individual components. 
VLM inference (5.40 s) and LLM reflection (5.75 s) dominate per-call computation time. 
However, these costly calls are not made at every step. 
The replanning interval ($N_{replan}=3$) reduces VLM decision calls by a factor of three compared to step-wise baselines. 
LLM reflection is triggered only upon stagnation detection, which occurs in 1.9 episodes on average (Table~\ref{tab:query_efficiency}). 
In contrast, spatial memory construction (D1) and heuristic planning (D2) operate at a millisecond scale, enabling continuous local movement without waiting for model responses. 
This hierarchical design effectively amortizes the high latency of foundation models across multiple steps and rare reflection events.

\begin{table}[h]
    \centering
    \caption{Average VLM queries per episode on GOAT-Bench (evaluated on RTX 3080 GPU).}
    \label{tab:query_efficiency}
    \setlength{\tabcolsep}{4.5pt} %
    \small
    \begin{tabular}{l||ccc|c}
        \toprule
        \textbf{Method} & \textbf{Base Nav.} & \textbf{Reflect.} & \textbf{Summary} & \textbf{Total $\downarrow$} \\
        \midrule
        3D-Mem & 31.6 & N/A & N/A & 31.6 \\
        \textbf{\sysname (Ours)} & \textbf{21.3} & 1.9 & 1.8 & \textbf{25.1} \\
        \bottomrule
    \end{tabular}
\end{table}

\textbf{VLM Query Efficiency.} 
Table~\ref{tab:query_efficiency} compares VLM query counts between \sysname and 3D-Mem on GOAT-Bench. 
3D-Mem queries the VLM at every step, averaging 31.6 calls per episode. 
\sysname reduces base navigation queries to 21.3 per episode through the replanning interval ($N_{replan}=3$), which commits the agent to a selected waypoint for multiple steps without re-querying. 
Additional calls for reflection (1.9) and summarization (1.8) are incurred only when stagnation is detected or memory reaches capacity. 
The total of 25.1 queries represents a 20.7\% reduction compared to 3D-Mem, demonstrating that decoupling semantic evaluation from spatial execution reduces computational cost without sacrificing navigation performance.

\subsection{Qualitative Results}

Figure~\ref{fig:qualitative} compares trajectories of \sysname and the baseline 3D-Mem across four goal modalities. 
Across all cases, 3D-Mem (red) is more easily distracted by locally plausible but globally incorrect cues, which leads to repeated turning, back-and-forth oscillation, and long detours before termination. 
For object and question goals, both methods can eventually reach the correct target, but \sysname follows noticeably shorter and smoother routes, indicating better path efficiency and less redundant revisiting. 
The advantage becomes more pronounced for image and description goals, where 3D-Mem is distracted by visually or semantically similar but incorrect regions and finally terminates at the wrong target. 
In contrast, \sysname (green) maintains smoother and more directed trajectories: the history-aware penalty suppresses revisiting recently explored dead ends, while metacognitive reflection helps the agent revise its search direction once exploration gain stalls. 
These examples illustrate that \sysname produces more efficient and robust trajectories under local visual or semantic ambiguity.

\section{Conclusion}\label{sec:conclusion}
\begin{sloppypar}
We presented \sysname, a training-free navigation framework that addresses local oscillation and redundant revisiting in vision-language navigation. 
The system integrates three key designs: Spatial Memory Construction builds a persistent 3D semantic map; History-Aware Heuristic Planning decouples semantic evaluation from spatial execution and suppresses redundant revisiting via an episodic penalty; Reflection and Correction detects stagnation and injects LLM-generated corrective rules back into the planner. 
Experiments on GOAT-Bench, HM3D-OVON, and A-EQA show that \sysname achieves state-of-the-art performance among training-free methods, with ablation studies confirming the contribution of each design.
\sysname offers a unified framework for learning from past failures and adapting future behavior, a step toward robust and self-correcting embodied agents.
\end{sloppypar}

\balance
\bibliographystyle{ACM-Reference-Format}
\bibliography{example_paper}

\end{document}